\documentclass{article}

 \usepackage[preprint]{neurips_2026}

\usepackage[utf8]{inputenc} % allow utf-8 input
\usepackage[T1]{fontenc}    % use 8-bit T1 fonts
\usepackage{hyperref}       % hyperlinks
\usepackage{enumitem}
\usepackage{xurl}            % simple URL typesetting
\usepackage{booktabs}       % professional-quality tables
\usepackage{amsfonts}       % blackboard math symbols
\usepackage{nicefrac}       % compact symbols for 1/2, etc.
\usepackage{microtype}      % microtypography
\usepackage{xcolor}         % colors
\usepackage{graphicx}
\usepackage{amsmath}

\hypersetup{pdfborder={0 0 0}} % Hides boxes; keeps it clean for print
\setlist[itemize]{leftmargin=*}
\setlist[enumerate]{leftmargin=*}

\title{LeCuration: A Tiny World Model as a Data Curation Multi-Tool}
\workshoptitle{Robot Learning with World Models: Capabilities, Frontiers, and Challenges}

\author{%
  Mayank Sengupta \quad Nirmit Desai\thanks{Corresponding author: \texttt{nirmit.desai@aintropy.ai}.} \quad Eric Song \quad Kunal Sawarkar \\
  AIntropy AI \\
}

\begin{document}

\maketitle

%------------------------------------------------------------------

\begin{abstract}
    Many applications of physical AI run within finite or closed physical worlds with a limited set of physical laws governing object behavior. Examples include robots working in a warehouse and agents moving around in a video game. In order to better organize, filter, and curate data for physical AI applications, we propose a new approach centered on the unique settings and physical laws of individual datasets. We train LeCuration, a small world model intended to serve as a \emph{data curation tool for a separate, larger downstream model}. To build this model, we choose LeWorldModel (LeWM) \citep{lewm} as our latent encoder and predictor, adding a diffusion transformer (DiT) decoder to add visuals to autoregressive gameplay rollout. We find that the embeddings of this model can be used as an anomaly detection signal and as a content-based clustering heuristic, and that auto-regressively predicting the game state with this model allows us to qualitatively check for action-state consistency. This paper presents a qualitative, proof-of-concept case study on CS:GO gameplay data; we do not yet report quantitative curation metrics or downstream training results, which we identify as the key next step.
\end{abstract}

%------------------------------------------------------------------

\section{Introduction}

Intelligently curating data prior to training is key to ensuring a high quality model, especially for use cases in physical AI. The straightforward alternative, feeding a model everything, teaches it the full distribution of the source data, including its degenerate corners. Filtering by hand does not scale. Rule-based filters require access to the physics engine state, which is hidden in most cases or impossible to access if your physics engine is the real world. Most importantly, it is impossible to understand the effect of a curated dataset without training a model on it. Therefore, we need a small, quickly trainable curation that has learned enough about a dataset to characterize it, identifying what is typical, what is rare, and what is surprising.

We train such a model on Counter-Strike: Global Offensive (CS:GO) gameplay recordings. LeCuration is a deliberately small model: the encoder has roughly 2M parameters, the predictor roughly 13M. By training it on this dataset in a matter of hours, we allow the model to produce representations and predictions that are semantically meaningful enough to drive curation decisions at scale. On top of this, we built a diffusion decoder as a visualization tool, letting us decode embeddings back to pixels to inspect what the model has learned.

Concretely, this paper's contributions are: (1) an adaptation of LeWM into a compact, quickly-trained curation tool operating on a TAESD latent space (Sec.~\ref{sec:arch}); (2) a qualitative demonstration that its embeddings organize gameplay footage into semantically coherent clusters usable for rebalancing and degenerate-content filtering (Sec.~\ref{sec:embeddings}); and (3) an unsupervised anomaly signal, derived from autoregressive embedding divergence, that flags frames where the predictor's expectation and the observed frame disagree (Sec.~\ref{sec:conclusion}). We emphasize that this is a proof-of-concept study: we validate these primitives qualitatively on a single CS:GO map, and we have not yet trained a downstream model on LeCuration-curated data or established quantitative metrics for any of the three primitives above; we discuss this scope explicitly in Sec.~\ref{sec:discussion}.

%------------------------------------------------------------------

\subsection{Background and Prior Work}
\label{sec:lewm}

The JEPA model training paradigm has been applied almost exclusively to two settings. 
\begin{itemize}
  \item The first is \emph{visual control and robot planning}: V-JEPA~2 \citep{vjepa2} trains on over one million hours of internet video to enable zero-shot robot manipulation by planning in latent space; Causal-JEPA \citep{causal_jepa} introduces object-level latent interventions to learn causal world structure; JEPA-based LiDAR models~\citep{lidar_jepa} apply the same framework to spatiotemporal occupancy forecasting for autonomous driving. 
  \item The second is \emph{representation learning for downstream classification and video understanding}: I-JEPA, V-JEPA, and VL-JEPA all frame prediction as a pretext task for learning transferable visual features.
\end{itemize}
In every case, the JEPA model is the \emph{end consumer} of its own representations. The learned embeddings and predictions are used either to plan the model's own actions or to evaluate the model's own performance on a benchmark task. In contrast, this paper investigates the application of the learned representations for data curation tasks.

\subsection{Main Contribution}

The inspiration of LeCuration is LeWorldModel (LeWM)~\citep{lewm}. LeWM's central contribution is a stable end-to-end JEPA that trains directly from raw pixels using only two loss terms: a next-embedding prediction loss and a Sketch Isotropic Gaussian Regularizer (SIGReg) that prevents representation collapse without stop-gradients, frozen encoders, or exponential moving averages. Much like the rest of the works in the space, LeWM is presented entirely as a \emph{planning and control} model. Given a goal frame, the predictor optimizes a sequence of actions whose predicted latent trajectory minimizes distance to the goal embedding.

We decided to build on this architecture for three main reasons: 

\begin{enumerate}
    \item At roughly 15M parameters, it is among the most compact world models that have demonstrated competitive performance on continuous control benchmarks. The smaller the model, the faster it trains, and the more practical it is to adapt it to new datasets if it is deployed in production.

    \item The authors establish the model's rich latent-space properties in their ablations. Physical quantity probing shows that agent position, object location, and object orientation are linearly recoverable from the CLS embedding (which is trained to be the overall per-frame world state). Surprise evaluation shows that the model assigns higher prediction error to physically implausible events (teleporting objects) than to matched visual perturbations that do not violate physics. Additionally, temporal path straightening shows that latent trajectories become increasingly linear during training, a property the authors argue reflects the emergence of a smooth, locally Euclidean world model.

    \item LeWM is not just a single-frame encoder model. While each frame is indeed individually encoded to the latent space, the predictor module of LeWM actually uses a history of $6$ previous vectors before predicting the next latent world state. This allows the model to understand the kinematics of domain-specific data, predicting world state not just from position, but also from velocity.
\end{enumerate}

We modify LeWM for a fundamentally different use: as a \emph{data curation tool for a separate, larger downstream model}, rather than as a planner. LeCuration does not plan actions and is not evaluated on any control benchmark. Instead, its embeddings and predictor outputs are used to analyze the training corpora.

Through this adaptation, we uncover several interesting properties. By clustering the embeddings, we find that we can qualitatively separate training data into visually and semantically distinct groups, such as frames from different map locations and camera framings (Sec.~\ref{sec:embeddings}); we have not measured this against ground-truth location or orientation labels. We can also run the predictor module autoregressively and compare the predictions with the encoded ground truth to find any possible anomalies or unexpected shifts in data. In addition, we build a diffusion transformer (DiT) decoder to recover gameplay frames conditioned on autoregressive CLS tokens output by the predictor module. By feeding a starting frame and a sequence of actions (key pushes) into the model, we are able to reconstruct near-photorealistic gameplay video from only a sequence of 196-dimensional world-state vectors, showing that despite its small size and very quick training time, LeCuration can indeed model the laws of a physics engine.

%------------------------------------------------------------------

\section{Architecture and Training}
\label{sec:arch}

\begin{figure}
    \includegraphics[width=1\textwidth]{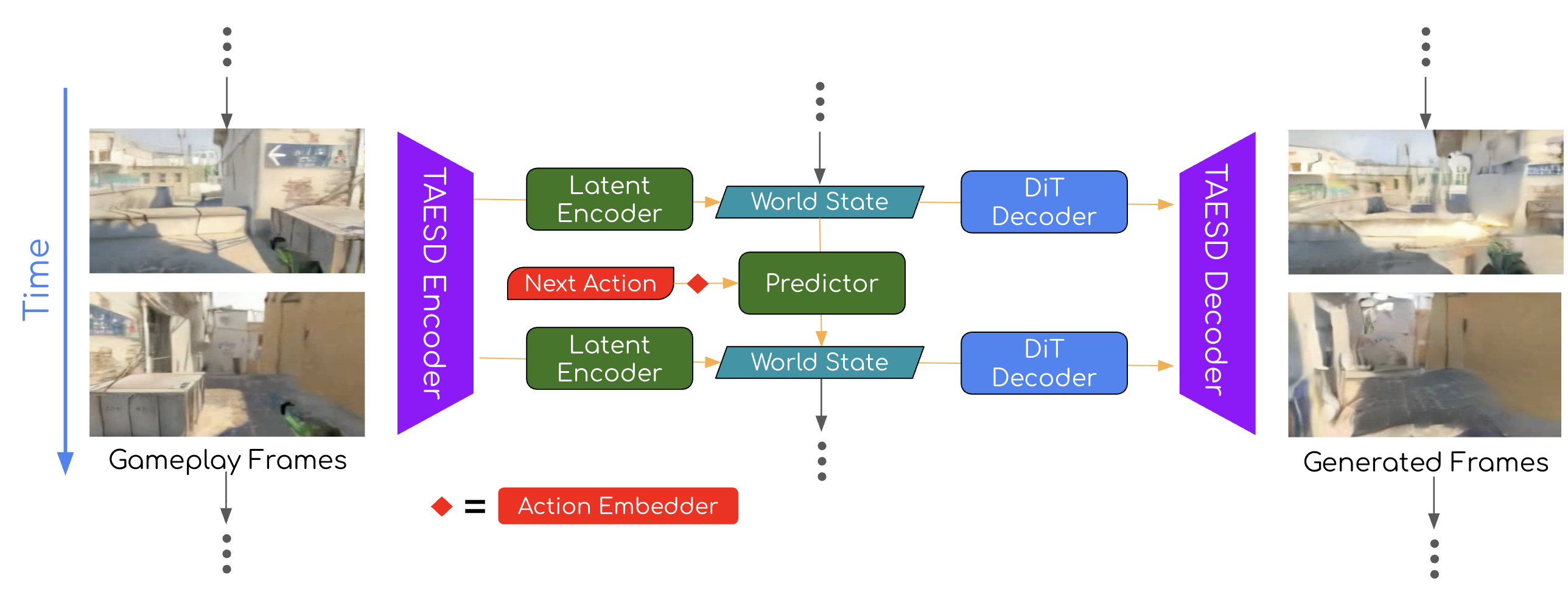}
    \centering
    \caption{The full architecture of LeCuration. Frames are compressed to StableDiffusion latents using TAESD, which are then encoded as world state vectors by the latent encoder. The latent predictor takes 6 past world states and the next action vector to predict the next world state, which is supervised by encoding the next ground truth frame as a world state. The Diffusion Decoder generates TAESD latents for each gameplay frame individually, conditioned only on the predicted world state. These are decoded to RGB frames by the TAESD decoder.}
    \label{architecture}
\end{figure}

While we take inspiration from both the architecture and training recipe of LeWM, we make major modifications, as shown in Figure \ref{architecture}, to improve training speed and fit our goals of applying it to data curation problems. 

\subsection{Stable Diffusion/TAESD Latent Space}
LeWM encodes directly from pixels. We decide to instead map our entire dataset to a much smaller intermediate latent representation before any training. We choose to use the same autoencoder as Stable Diffusion \citep{stablediff}, as it has been standard practice, and is able to do $48\times$ compression while losing relatively few important details. However, the autoencoder used by Stable Diffusion has $74$M parameters, and a much more efficient alternative, called Tiny AutoEncoder for Stable Diffusion (TAESD) \citep{taesd}, accomplishes the exact same task at a fraction of the cost. Distilled from the standard Stable Diffusion autoencoder, TAESD only has $2.4$ M parameters and a single forward pass is roughly $20\times$ faster than that of the original autoencoder on the same device \citep{taesd}. 

Our dataset is thus compressed from $224\times224\times3$ frames to $4{\times}28{\times}28$ TAESD latents. Each frame has a corresponding action vector, which is a $51$-dimensional one-hot encoded vector representing the keys pressed at that frame (key presses, mouse movements, and so on).

\subsection{Latent Encoder and Predictor}

The latent encoder and predictor were inspired by LeWM but adapted to the TAESD latent space. The Latent Encoder is a patch-ViT operating on TAESD latents (patch size $4$, $49$ spatial patches plus a CLS token [the world state], projected to $196$ dimensions via $4$ transformer layers). For each gameplay frame, it produces a single $196$-dimensional world state $e_{t+1}$ (the CLS token) that summarizes the frame's content. The patch tokens are discarded after each forward pass. 

The latent predictor is a causal transformer ($512$ hidden dimension, $8$ layers, $16$ heads) that takes a sliding window history of $6$ frame embeddings plus action embeddings and predicts the next frame's embedding. Actions are projected to $196$ dimensions by a small action embedder before being fed to the predictor. The action embedder is a 1d-convolution (kernel\_size=10, stride=1) followed by a 2-layer MLP (hidden\_dim=768) which transforms the 51-dimensional one-hot encoded action vector into a 196-dim embedding.

At each autoregressive step the predictor outputs world state $\hat{e}_{t+1}$: its expectation of what the next frame should look like in embedding space, given the recent context and the action taken. The quality of this prediction---and in particular the gap between $\hat{e}_{t+1}$ and the observed $e_{t+1}$---is the primary curation signal.

\subsection{Diffusion Transformer Decoder}

A Diffusion Transformer (DiT) Decoder ($512$ hidden dimension, $8$ layers, CFG dropout
$0.1$) is trained to denoise TAESD latents conditioned on the frozen CLS
embedding. This lets us decode any embedding back to pixels to visually
inspect what the encoder has captured. All figures showing rendered frames
use this decoder. It is trained after JEPA pre-training and plays no role in
the curation pipeline.

\subsection{Training Recipe}

We pre-encode and shard all frames in the dataset as TAESD latents. Performing this forward pass exactly once per frame not only saves redundant compute, but slashes I/O cost, as (1) $48\times$ less data is fetched per training step and (2) compressing the dataset allows us to fit all of the frames and corresponding actions on disk, so that we don't have to repeatedly fetch batches from our cloud bucket.

Like all JEPA architectures, LeCuration training is self-supervised: the predictor's output is compared against a stop-gradient target produced by a slow EMA copy of the encoder (EMA decay $0.9999$). For a frame at arbitrary time step $t$ in a gameplay sequence, the loss is calculated as:
\begin{align}
    \mathcal{L} = \text{MSE}(e_{t-6 : t}, \hat{e}_{t-6 : t}) + \lambda \cdot \text{SigReg}(e_t)
\end{align}
where $e_{t-6 : t}$ and $\hat{e}_{t-6 : t}$ are the world states embedded from the ground truth history and next frames and the world states for the same frames as predicted by the latent predictor, respectively. SIGReg is short for Sketch Isotropic Gaussian Regularizer, which prevents representation collapse by pushing the embedding distribution toward an isotropic Gaussian. We use the regularization coefficient $\lambda = 0.09$ as recommended by the LeWM authors. No labels, no collision annotations, no physics supervision---just gameplay video and actions.

There are two stages of training for the latent encoder and predictor. 
\begin{itemize}
    \item First, we pre-train the encoder on randomly sampled frame-action pair sequences, asking the encoder to find the world state of $6$ history frames and the predictor to predict only the next immediate world state. This allows the encoder to form initial representations for the training data. 
    \item Next, we do supervised auto-regressive post-training, where, starting from a window of $6$ frames, we ask the predictor to auto-regressively roll out word states for the next $50$ frames, comparing them with the world states encoded directly from the ground truth frames. We use a discount factor of $1.03$ to emphasize frames further away. This significantly improves the quality of longer rollouts from the predictor, and also allows the encoder to refine its world state embeddings. 
\end{itemize}

No part of the model is frozen in either of these two stages. Pre-training takes \textbf{120 minutes} for $300$ epochs, and post-training takes \textbf{60 minutes} for $20$ epochs. We define an epoch to be 1600 batches, rather than a full pass through the entire dataset. Pre-training batch size is $32$ and post-training is $16$. A batch denotes $7$ frame-action pairs in pre-training ($6$ history frames + $1$ current frame) and $56$ frame-action pairs in post-training.

For the DiT decoder, we embedded TAESD latents from the dataset as world states with the trained latent encoder, and trained the decoder to generate the TAESD latents conditioned on each world state. This way, the DiT decoder performance is unable to affect the world model quality. This step was also very fast (30 mins for 600 epochs). 

\subsection{Dataset and Compute}

The dataset in our experiments in this paper is composed of CS:GO gameplay recordings of various skill levels restricted exclusively to the \texttt{de\_dust2} map. The data was provided anonymously by our data partner. Both frames and key presses are sampled at $16$Hz. No curation or episode filtering is applied at training time; the model can be used for curation applications after being trained on the full, uncurated set. The idea is to train LeCuration on any training corpus and then apply it to curate sample-efficient training sets.

All experiments were conducted on a runtime with a 40GB Nvidia A100 GPU accessed through Google Cloud Platform (GCP). Our data was also stored in GCP buckets.

%------------------------------------------------------------------

\section{Evaluation: Embedding and Predictor Quality}
\label{sec:embeddings}

We qualitatively evaluate the embedding and predictor outputs in the context of addressing the common data curation challenges.

\subsection{Sorting the Dataset by Clustering Embeddings}

We extracted one million CLS embeddings from the training corpus and ran UMAP + HDBSCAN to cluster them. The clusters correspond to recognizable gameplay situations---long open sightlines, close-quarter interiors, smoke-obscured views, player silhouettes at doorframes---without any semantic labels having been provided. The self-supervised prediction objective has organized the space by content, not by pixel statistics.

\begin{figure}[h]
\centering
\includegraphics[width=0.85\textwidth]{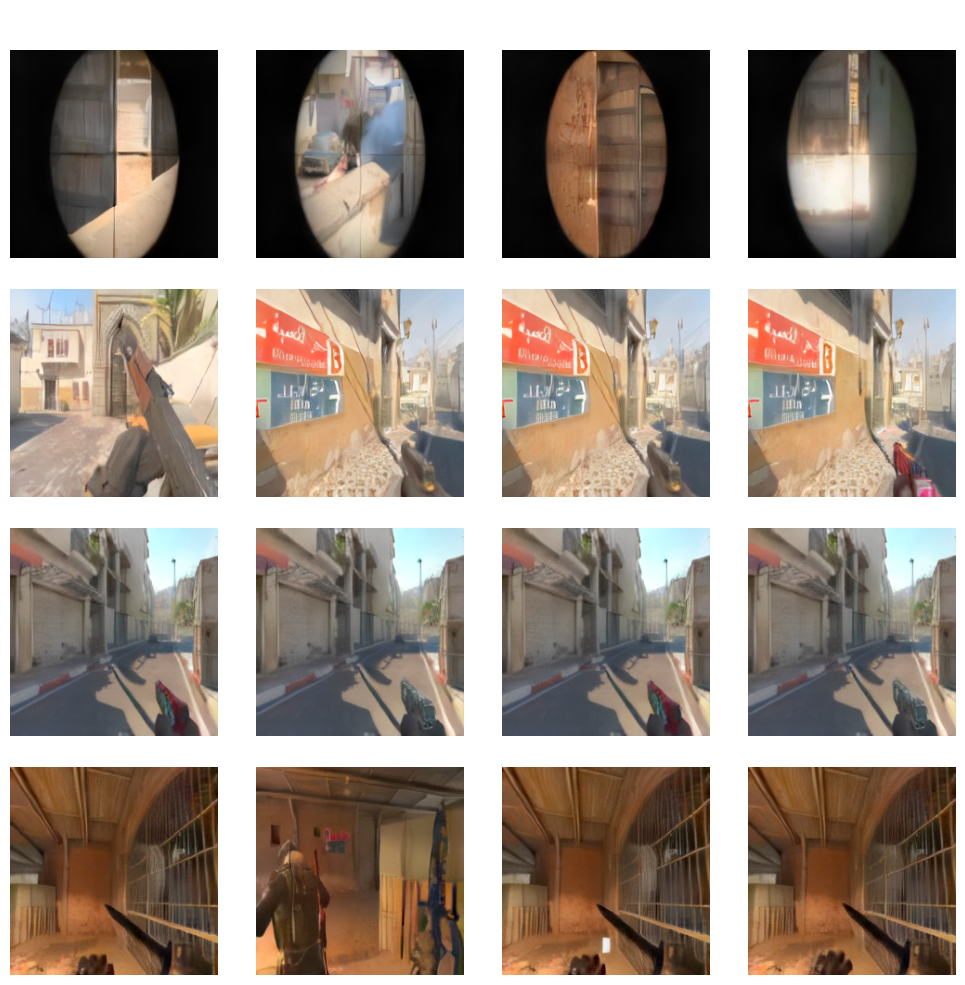}
\caption{Frames sampled from the dataset after HDBSCAN clustering. Each row is one cluster of world states in the latent space. Visual coherence within rows and clear differences between rows confirm that the learned embeddings recover semantically meaningful groupings of gameplay footage without any manual annotation. Additionally, the first row demonstrates how we can sift out anomalous clusters (in-scope views) without having to search for them specifically.}
\label{clusters}
\end{figure}

This cluster structure provides direct curation primitives. \textbf{Rebalancing}: over-represented clusters (common corridor walks, static standoffs) can be downsampled at episode selection time to improve training diversity. \textbf{Degenerate-content filtering}: clusters corresponding to loading screens, spectator views, and scope views (the first row of Figure \ref{clusters}) or heavily compressed frames can be easily filtered out, without per-frame labels.

\subsection{Auto-regressive Nearest-Neighbor Retrieval}
\label{sec:nnr}

To test whether the embedding geometry is meaningful independently of the generative decoder, we ran a nearest-neighbor retrieval experiment. At each step of an auto-regressive rollout, instead of rendering the predicted embedding with the DiT, we snap it to its nearest neighbor (by Euclidean distance) in the 1M-frame subset of our training data. We decode and append the corresponding TAESD latent through the VAE directly to generate the video. If the embedding space is coherent, the retrieved frame should be visually consistent with what the model predicted.

The retrieved frames match the ground-truth sequence in scene context, player orientation, and lighting---the predictor's output lands near real frames showing the right kind of gameplay situation \footnote{\url{https://drive.google.com/file/d/1p-f8nqGOt85z5uXaiNs9Cu-JBPv1rJ1l}}. This confirms that (1) the embedding geometry captures genuine semantics, and (2) a 1M-frame database is dense enough to serve as a retrieval corpus. It also suggests a lightweight curation primitive: selecting training frames by proximity to a target embedding, without running the DiT at all.

Studying the world states with nearest-neighbor retrieval allowed us to find some of its weaknesses. A major weakness is that the weapon in the player's hand is often misrepresented. This can be seen in the third row of Figure \ref{clusters}, where the location and camera angles of the clustered frames are obviously the same, but the weapon is not. When playing an auto-regressively predicted sequence of nearest-neighbor snapped frames, the weapon type often flickers in the player's hand. This is likely because the weapon takes up significantly less space on the screen than the surrounding scene, meaning it has a much smaller effect on the final loss value. Also, the action inputs that involve the weapon, firing, and switching, are much less common than moving around and turning. This is the kind of result that proves the usefulness of training a small scout model. If we were to train a larger model afterwards, we would know to create a curated finetuning set that oversamples weapon switches and firing to improve accuracy in this aspect. 

\subsection{Using Clusters for Coverage-Based Curation}

We used the per-episode embedding centroids to select training episodes by
farthest-first traversal (maximizing spread in the 196-dimensional space).
By construction, coverage-maximizing selection produces a more diverse
training corpus with less redundancy than random sampling. We have not yet
trained a downstream world model on these subsets, so we do not yet know
whether this translates into comparable downstream quality at a reduced
training budget; validating this is the most important open question raised
by this work (see Sec.~\ref{sec:discussion}).

Rare-trajectory upsampling selects episodes occupying sparse embedding regions
(high average distance to $k$-nearest neighbors). These frames represent
situations the model has seen few examples of, and we hypothesize that they
would carry disproportionate training signal for a downstream model if
upsampled; we have not yet tested this hypothesis. We plan to share the
results of training a larger world model on coverage- and rarity-curated
subsets, compared against random-sampling and full-corpus baselines, in
future work.

%------------------------------------------------------------------

\section{Predictor Quality and Anomaly Detection via Embedding Divergence}
\label{sec:conclusion}

\begin{figure}[h]
\centering
\includegraphics[width=0.85\textwidth]{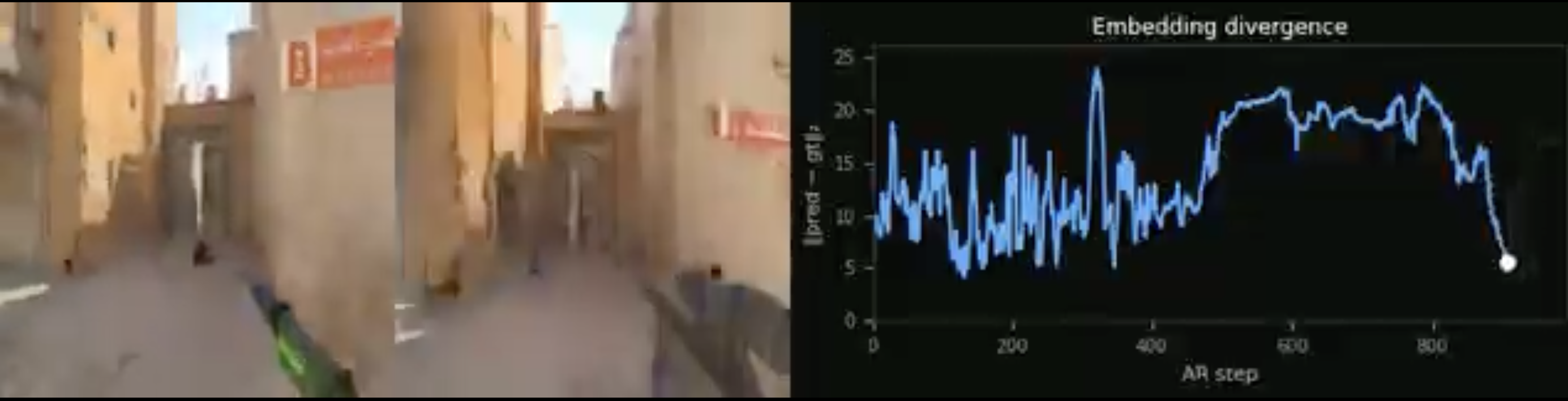}
\caption[Animated embedding divergence during autoregressive rollout.]%
{Animated embedding divergence during autoregressive rollout. \textbf{Left}: ground-truth CS:GO frame. \textbf{Center}: Each predicted world state decoded back to an RGB frame using the DiT. \textbf{Right}: Euclidean distance $d_t = \protect\|\hat{e}_t - e_t\protect\|_2$. Spikes identify frames where the predictor's expectation diverges from reality, providing an unsupervised anomaly signal derived entirely from the world model's own predictions.}
\label{divergence}
\end{figure}

The predictor's usefulness for curation goes beyond producing embeddings for
retrieval. The video in Figure \ref{divergence} is created by starting with a queue of $6$ ground truth frames, encoding them with the latent encoder, and auto-regressively generating world states until the end of the ground truth clip. Each world state is decoded by the DiT to generate the video. During auto-regressive rollout, we track the Euclidean distance between the predictor's output and the world state encoded directly from the ground-truth frame at each step:
\[
  d_t = \|\hat{e}_t - e_t\|_2
\]
where $\hat{e}_t$ is the predictor's forecast and $e_t$ is the world state
of the actual observed frame at step~$t$. 

When the rollout proceeds normally---predictable camera motion, standard
player movement---$d_t$ stays low. When the rollout reaches a frame that
differs substantially from what the predictor expected, $d_t$ spikes. We
interpret these spikes as candidate moments of surprise: abrupt events, unusual
trajectories, or physical interactions outside the model's learned distribution.
We note that a spike is not on its own diagnostic of the cause: it can also
arise from accumulated rollout error or from out-of-distribution actions, and
we have not yet defined a quantitative threshold or metric for distinguishing
these cases or for selecting which high-$d_t$ frames to discard versus
oversample. We treat $d_t$ as a candidate signal for further investigation
rather than a validated curation rule, and discuss this limitation further in
Sec.~\ref{sec:discussion}.

Figure \ref{divergence} shows how the latent predictor's world states diverge when a rare or confusing set of actions is taken \footnote{\url{https://drive.google.com/file/d/1vbJ9v4vKoFvI21PR7D1jFkBlQbWtAGWc}}. The scene teleports to a different part of the map, but snaps back to the correct location after about 350 frames. This period of divergence is clearly visible on the graph in Figure \ref{divergence}.

This anomaly signal is entirely learned. It emerges purely from the gap between predicted and observed embeddings. Frames with high $d_t$ are candidates for anomaly-based curation: either to be excluded (if they represent degenerate gameplay) or oversampled (if they represent rare but physically meaningful interactions that a more robust downstream model needs to capture accurately). 

The predictive world model is itself the detector: anything that surprises the predictor registers as high $d_t$. The current limitation is that the signal is not physics-specific. A player taking an unusual path, an abrupt cut in the recording, and a genuine wall-collision event all produce $d_t$ spikes. Isolating physics-caused divergence from other causes---such as eliminating scenes with double jumps---would require a model pretrained on a wide corpus of real-world data, not just gameplay data, and is a direction for future work.

%------------------------------------------------------------------

\section{Discussion and Limitations}
\label{sec:discussion}

The central claim of this work is that a small domain-trained predictive world model is a useful tool for data curation. The encoder (2M parameters) and
predictor (13M parameters) together represent a tiny fraction of the downstream
DIAMOND-style \citep{alonso2024diffusion} world model they are curating data for. Yet they provide:

\begin{itemize}
  \item A semantically organized embedding space that clusters gameplay by
    content without labels.
  \item A qualitative signal of which aspects of the data need to be oversampled for a larger downstream model.
  \item A learned, dataset-specific anomaly signal from embedding divergence during autoregressive rollout.
\end{itemize}

The cost of running this curation model over the full training corpus is
negligible compared to training the downstream world model. Embedding
extraction is a single forward pass per frame; anomaly scoring is an
additional forward pass of the predictor. Both can be cached and reused across
any number of downstream curation experiments without retraining.

The main limitation is physics specificity: the embeddings and predictions
encode appearance and motion continuity, not physical plausibility. Closing
that gap---moving from a model that detects what is \emph{visually surprising}
to one that detects what is \emph{physically implausible}---is a natural next
step and the subject of ongoing work.

Additionally, we want to test the capabilities of this model on a larger and more varied set, preferably from real world data rather than gameplay. An ideal dataset would be robotic manipulation corpora, which have video and exact robot actions for a wide variety of tasks and settings. This would present a much greater challenge to the LeCuration model and show its full potential for data curation in the physical AI space. We are confident that our approach, once refined, can introduce a paradigm shift in the sourcing and curating of data for all types of models, and we are excited to see where the industry takes it from here.

%%%%%%%%%%%%%%%%%%%%%%%%%%%%%%%%%%%%%%%%%%%%%%%%%%%%%%%%%%%%

\appendix

\section{Comparing Nearest-Neighbor Retrieval and DiT Output}

The DiT was designed to be completely separate from the core world model, namely the latent encoder and predictor, and serve only as a visualization tool. We found that a DiT was not strictly necessary as a visualization tool once the model saw enough of our data. At that point, we found that we could randomly sample $1$ million frames from our dataset and create a nearest neighbor retrieval index from their world state embeddings. By snapping autoregressively generated world states with their nearest neighbor as we did in Section \ref{sec:nnr}, we were able to create more interpretable roll out videos (mainly because the real frames fetched from the dataset were clearer than those generated by the DiT).

This allowed us to better understand some of the divergence points. For example, if the ground truth video showed a fellow player passing by and obscuring a large section of the screen, this could cause a divergence. The DiT would not be able to clearly depict the passing player, as such scenes were rare in its training data. 

However, for future work, we would recommend using both a DiT and nearest-neighbor retrieval to understand generated video. Nearest-neighbor retrieval only worked well for this dataset because all the scenes come from the same relatively small and closed map (\texttt{de\_dust2}). It may still help make interpretable generated videos in other cases, but with a more varied dataset a DiT would be helpful in decoding and interpolating world states that represent richer data. 

\section{Code and Data}
\label{sec:code}

\begin{itemize}
  \item All code for this project is available at: \url{https://drive.google.com/file/d/1YDHDfSfwi6VOp4wXA8OXajeyIgH5G39u}. \item The exact dataset we used is not available to the public, as it was provided to us by an anonymous data partner. However, a very similar Counter-Strike dataset (from the game Counter-Strike 2) in the exact same format as our dataset is public: \url{https://huggingface.co/datasets/TeaPearce/CounterStrike\_Deathmatch}.
\end{itemize}

%%%%%%%%%%%%%%%%%%%%%%%%%%%%%%%%%%%%%%%%%%%%%%%%%%%%%%%%%%%%

%\newpage
% PUT THIS BACK LATER
%\input{checklist.tex}

\end{document}